\documentclass[review]{elsarticle}
\usepackage{setspace}  
\usepackage[numbers]{natbib}
\usepackage{amsmath,amssymb,amsfonts}
\usepackage{algorithmic}
\usepackage{graphicx}
\usepackage{textcomp}
\usepackage{epstopdf}
\usepackage{subfig}
\usepackage{url}
\usepackage{hyperref}
\usepackage[linesnumbered,ruled]{algorithm2e}
\usepackage{xcolor}
\usepackage{colortbl, booktabs} 
\usepackage{framed,multirow}
\usepackage{float}
\usepackage{amssymb}

\journal{Neurocomputing}

\begin{document}

\begin{frontmatter}

\title{FU-Mamba: A Frequency-Enhanced Dynamic Scanning Framework for Oralscan Image Segmentation} 

\author[1]{Xinxin Zhao}
\author[1]{Jinpeng Ye}
\author[1]{Bo Wei}
\author[2]{Liqin Wu}
\author[3,10]{Mahmoud Hassaballah}
\author[4]{\\Karen Egiazarian}
\author[5]{Aura Conci}
\author[6]{Victor Hugo C. de Albuquerque}
\author[7]{\\Abdulkadir Sengur}
\author[8,9]{Leszek Rutkowski}
\author[1]{Yan Tian\corref{cor1}}
\ead{tianyan@zjgsu.edu.cn}

\affiliation[1]{organization={School of Computer Science and Technology, Zhejiang Gongshang University},
                postcode={310018}, 
                city={Hangzhou},
                country={China}}

\affiliation[2]{organization={Department of Stomatology, Tongxiang Hospital of Traditional Chinese Medicine},
                postcode={314500}, 
                city={Tongxiang},
                country={China}}

\affiliation[3]{organization={Department of Computer Science, Prince Sattam Bin Abdulaziz University},
                postcode={16278}, 
                city={AlKharj},
                country={Saudi Arabia}}

\affiliation[4]{organization={Department of Computing Sciences, Tampere University},
                postcode={33720}, 
                city={Tampere},
                country={Finland}}

\affiliation[5]{organization={Department of Computer Science, Universidade Federal Fluminense},
                postcode={24210-346}, 
                city={Niteroi},
                country={Brazil}}

\affiliation[6]{organization={Department of Teleinformatics Engineering, Federal University of Ceara},
                postcode={60020-181}, 
                city={Fortaleza},
                country={Brazil}}

\affiliation[7]{organization={Department of Electrical and Electronic Engineering, Faculty of Technology, Firat University},
                postcode={23000}, 
                city={Elazig},
                country={Turkey}}

\affiliation[8]{organization={Systems Research Institute, Polish Academy of Sciences},
                 postcode={01-447},
                 city={Warsaw},
                 country={Poland}}

\affiliation[9]{organization={AGH University of Krakow},
                 postcode={30-059},
                 city={Krakow},
                 country={Poland}}

\affiliation[10]{organization={Department of Computer Science, Qena University},
                 postcode={83523},
                 city={Qena},
                 country={Egypt}}

\cortext[cor1]{Corresponding author}

\begin{abstract}
Oralscan image segmentation is essential for computer-aided diagnosis and treatment planning in digital dentistry. However, existing visual state space models (SSMs) often rely on manually designed scanning orders to flatten image patches into sequences, which disrupts the semantic spatial continuity and hinders coherent feature extraction from key foreground regions. Moreover, elements such as inconsistent lighting, reflective surfaces, and noise during data acquisition disrupt the frequency distribution by diminishing high-frequency details while enhancing low-frequency components, consequently hindering the accurate localization of boundaries. In response to these challenges, we introduce FU-Mamba, an innovative framework that incorporates dynamic scanning and frequency domain enhancement within the SSM architecture. Specifically, the Dynamic Mamba Block (DMB) adaptively learns sampling offsets via a trainable offset prediction network and performs flexible bilinear interpolation, enabling content-aware scanning that preserves spatial coherence. Furthermore, a frequency domain enhancement block balances spectral components through wavelet-guided decomposition and spectrum pooling, improving robustness under adverse imaging conditions. Experimental findings indicate that FU-Mamba attains a notable enhancement in segmentation accuracy, evidenced by a 1.1\% increase in the mean intersection over union (mIoU) metric when evaluated on the dental segmentation dataset. Project page: \href{https://byte2bite.github.io/FU-Mamba/}{ https://byte2bite.github.io/FU-Mamba/}
\end{abstract}


\end{frontmatter}


\section{Introduction}
\label{sec:introduction}
Tooth segmentation plays a vital role in digital dentistry, with potential applications in disease diagnosis~\citep{peng2026adml}, treatment monitoring~\citep{tian2023revised,tian2024rgb}, and image-based clinical analysis. Effective segmentation enhances the detection of dental conditions~\citep{azad2026unified}, including caries, periodontal disease, and various oral abnormalities, thus allowing for more prompt and accurate diagnoses.

\begin{figure}[htbp]
\centering
	\includegraphics[width=0.72\linewidth]{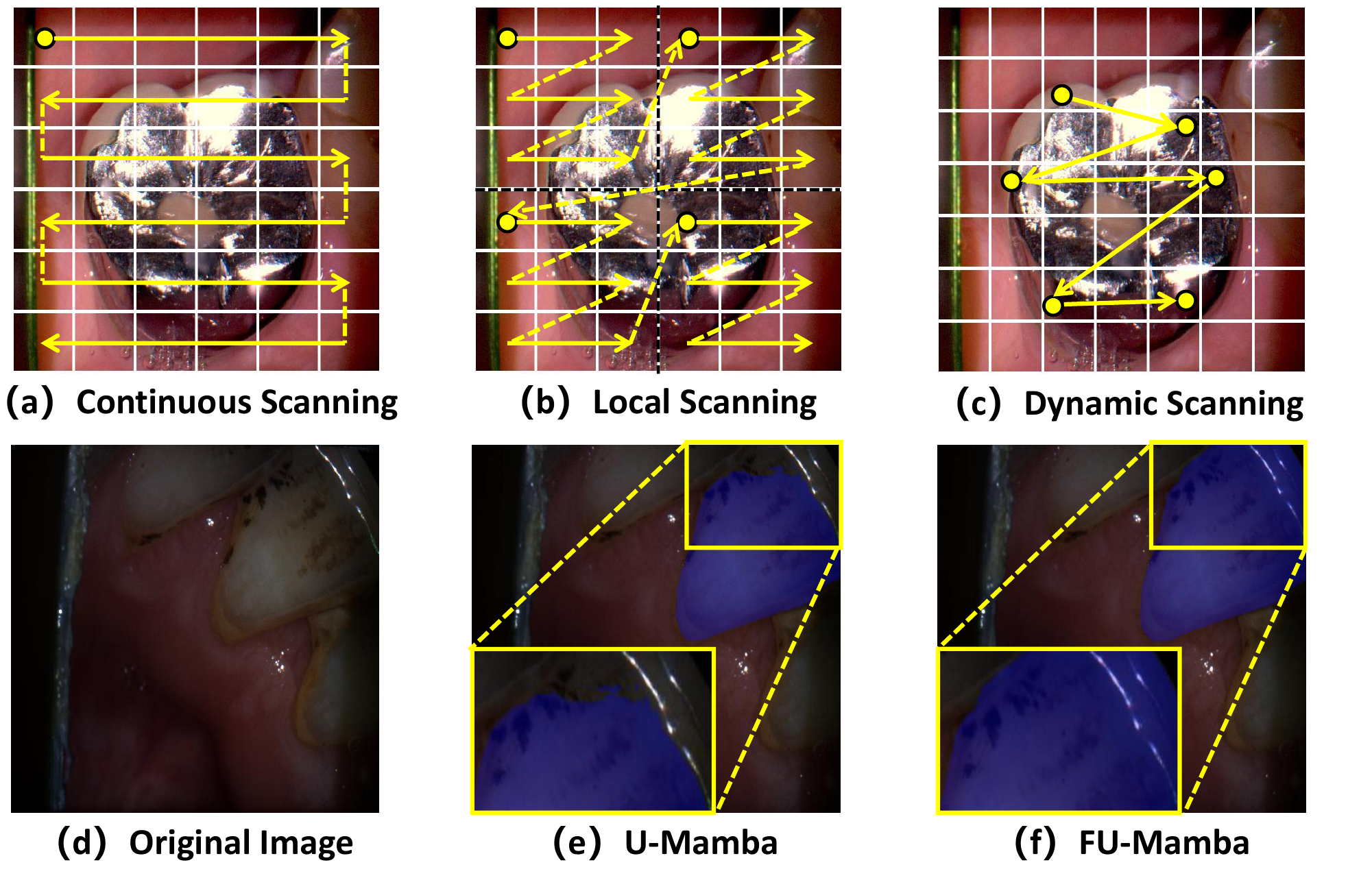} \\
	\caption{\textbf{Comparison of different scanning strategies in visual state space models.} (a) Continuous scanning strategy proposed by PlainMamba; (b) Local scanning strategy proposed by LocalMamba; (c) Our proposed dynamic scanning strategy; (d) Original dental image under poor lighting conditions; (e) Low-quality segmentation mask produced by U-Mamba; (f) Segmentation mask generated by our approach. Yellow dots indicate reference points, and yellow arrows represent the scanning order. 
	}\label{figchallenges}
\end{figure}

State space models (SSMs)~\citep{SSM}, as exemplified by Mamba~\citep{Mamba}, have garnered significant interest in recent times due to their ability to selectively retain or discard information across sequences via the S6 block while offering both linear-time computation and a global receptive field. Inspired by Mamba’s success in natural language processing, several works, such as Vim~\citep{VisionMamba}, have broadened its applicability to the field of computer vision by transforming two-dimensional images into one-dimensional sequences through the implementation of diverse scanning methodologies. This approach illustrates the capabilities of SSMs in visual tasks.
As demonstrated in Fig.~\ref{figchallenges}(a), PlainMamba~\citep{PlainMamba} contends that traditional raster scanning fails to account for the significance of spatial continuity in images. Consequently, it proposes a continuous scanning approach aimed at maintaining the correlation between neighboring patches. Likewise, as shown in Fig.~\ref{figchallenges}(b), LocalMamba~\citep{huang2024localmamba} proposes a local scanning strategy to better capture spatial positional relationships. Nevertheless, these scanning approaches disrupt the spatial adjacency of semantic structures after flattening~\citep{tian2026DM-CFO}, leading to the loss of critical structural information. By contrast, as shown in Fig.~\ref{figchallenges}(c), our approach adaptively determines the scanning order based on image content. While recent advances like DefMamba~\citep{liu2025defmamba} have explored deformable state spaces for natural images, our approach uniquely synergizes this content-aware dynamic scanning with frequency balancing, enabling more flexible modeling capabilities tailored for complex dental artifacts. To demonstrate our scanning approach, the figure displays six sample points, though the full process includes additional points.


Effectively segmenting images with complex and variable frequency characteristics remains a challenging task, particularly in dental image analysis, where the input is often affected by uneven lighting, reflective surfaces like enamel and metal restorations, motion blur, and noise during image capture. These artifacts modify the original frequency distribution of the image by attenuating high-frequency components, such as edges and textures, while redistributing the overall energy towards lower frequencies. Consequently, the model struggles to precisely detect subtle features like tooth edges and gum shapes.
As shown in Fig.~\ref{figchallenges}(e), U-Mamba fails to effectively balance high- and low-frequency components, resulting in inaccurate tooth boundary segmentation and an inability to completely segment the tooth under poor lighting conditions. In comparison, as demonstrated in Fig.~\ref{figchallenges}(f), our approach attains a more effective equilibrium among frequency components and successfully delineates the entire tooth region, even under less than ideal illumination conditions.

Our work is inspired by recent advances in visual state space modeling and frequency-aware representation learning. In computer vision applications, the Mamba approach breaks down 2D images into smaller patches. These patches are then converted into several 1D sequences by employing specific scanning techniques from different angles. This approach successfully integrates the Mamba model for visual applications, achieving strong results and demonstrating the promise of SSMs in computer vision.
In parallel, WTConv~\citep{WTConv} utilizes wavelet-based decomposition to improve the network's capacity to identify hierarchical spatial patterns. This approach is especially advantageous for the representation of large-scale structural information in natural images. SPAM~\citep{SPAM} introduces a frequency-aware convolutional mixer that adaptively integrates spectral cues into spatial representations, thereby strengthening feature discriminability.

To address the challenges of spatial structure disruption and frequency imbalance in dental image segmentation, we propose a novel framework named FU-Mamba, which integrates a dynamic scanning mechanism and frequency-domain enhancement within a state space model. Specifically, we introduce a DMB that adaptively learns sampling positions and scanning sequences during training. By dynamically adjusting the scanning path to prioritize important information, this block improves the acquisition and processing of pertinent input features. As a result, the model attains enhanced flexibility in its modeling capabilities while preserving linear computational complexity and a comprehensive global modeling capacity.
Additionally, to augment the model's proficiency in managing intricate frequency characteristics, we integrate a frequency domain enhancement block (FEB) specifically engineered to equilibrate various frequency components. This module integrates a wavelet-guided spectral pooling module (WSPM), which not only amplifies frequency-domain information but also emphasizes mid-frequency signals aligned with human visual sensitivity.
Fig.~\ref{method} presents a comprehensive depiction of the methodology we have proposed.

The key contributions of this study are delineated as follows:
\begin{itemize}
\item A unified approach named FU-Mamba is developed by integrating dynamic scanning and frequency-aware modeling into a visual state space model for accurate and efficient dental image segmentation.
\item A DMB is introduced to adaptively adjust sampling positions and scanning sequences based on input features, thereby preserving spatial continuity and enhancing structural representation.
\item A frequency domain enhancement block is proposed to balance high- and low-frequency components by integrating wavelet-based decomposition and spectrum pooling, improving robustness under challenging imaging conditions.
\end{itemize}

Experimental results on the dental segmentation dataset (DSD)~\citep{TIAN2020107158} and OralVision~\citep{oralvision} dataset demonstrate that the proposed FU-Mamba framework effectively generates high-quality segmentation masks.

The layout of the subsequent sections is as follows: Section \ref{relatedWork} offers a summary of studies focused on vision mamba and frequency domain analysis. Section \ref{ourApproach} outlines the proposed methodology in detail. Section \ref{results} presents the findings from the experiments conducted. Finally, Section \ref{conclusion} concludes the paper and explores possible avenues for future research.

\section{Related Work}\label{relatedWork}
We present a concise overview of the contemporary literature concerning dental image segmentation, vision mamba, and frequency domain analysis. Additionally, it offers a comparative analysis of the advantages and disadvantages associated with the current methodologies in these areas.

\subsection{Dental Image Segmentation}
Dental image segmentation~\citep{zhao2026innovative} represents a critical area of inquiry within the domain of computer vision, especially in the context of dental radiological diagnostics. The efficacy of segmentation algorithms is paramount for the precise identification of dental structures and the development of subsequent treatment strategies. In recent years, scholars have introduced a range of novel segmentation techniques~\citep{awari2024three,TIAN2020107158,tian20223d,tian2023revised,chen2024virtual,tian2024rgb} aimed at overcoming the challenges associated with dental image segmentation.

Du et al.~\citep{wenjie2024combined} introduced a hierarchical segmentation level set model that markedly enhances segmentation accuracy by mitigating the transfer and accumulation of segmentation errors. Their approach provides a promising foundation for further studies in this field. In the realm of deep learning, Joshi et al.~\citep{joshi2024segmentation} employed the UNet model to segment panoramic X-ray images, demonstrating the power of advanced machine learning in medical imaging. Furthermore, Harsh et al.~\citep{DentalUnet} extended this line of work by incorporating an attention module into the UNet framework and fine-tuning it for dental image analysis. The incorporation of an attention gate allows this approach to dynamically concentrate on critical feature regions within dental images while minimizing the influence of extraneous information, thus improving both segmentation accuracy and robustness.

Additionally, Hou et al.~\citep{hou2023teeth} introduced Teeth UNet, an enhanced UNet variant designed specifically for dental panoramic X-ray segmentation. This model enhances the delineation of tooth boundaries and contextual semantic information through the implementation of dense skip connections, multi-scale aggregation attention blocks, and dilated hybrid attention blocks, effectively addressing challenges such as indistinct tooth boundaries and low contrast. This model outperforms the standard UNet architecture by a significant margin, delivering sharper segmentation results and more precise feature detection, especially when dealing with intricate dental anatomy and images with poor contrast. In their quest for streamlined solutions, Lin et al.~\citep{lin2023lightweight} pioneered a lean deep learning framework tailored for segmenting dental X-ray images, making it viable for edge device implementation. By leveraging knowledge distillation techniques, they refined the neural network's design, reducing computational overhead and parameter counts without compromising accuracy. This breakthrough paves the way for real-time clinical use, striking an ideal balance between efficiency and performance.
However, despite the significant progress made by these approaches in dental image segmentation, their segmentation results still exhibit certain limitations when dealing with oral edge regions and under insufficient lighting conditions. Beyond traditional mechanisms, recent advancements have explored diverse paradigms. For instance, novel backbones like U-KAN~\citep{ukan} and Light-UNETR~\citep{ultralightvmunet} have demonstrated strong capabilities. Furthermore, interactive and ambiguity-aware models such as MedSAM-Agent~\citep{medsam_agent2026}, P2SAM~\citep{p2sam2024}, and others~\citep{flaws_applause2024} empower segmentation through reinforcement learning or probabilistic prompting. Innovations like GTP-4O~\citep{gtp4o2024} and global correlation networks~\citep{fewshot_correlation2020,liu2021part,liu2025part} further advance biomedical representation.

\subsection{Vision Mamba}
Mamba~\citep{Mamba} was originally introduced for natural language processing, where it demonstrated strong performance in sequence modeling by selectively propagating input-dependent information through an efficient state space mechanism. This success has motivated the application of Mamba to various visual tasks, including medical image segmentation~\citep{MedMamba,VM-Unet2}, remote sensing image segmentation~\citep{rs3mamba,PMamba,hou2025toward,hou2025fedc}, and object detection~\citep{mambayolo}. Unlike natural language sequences, images do not have a unique causal ordering, making the design of suitable scanning strategies crucial for visual Mamba models. To address this issue, Vim~\citep{VisionMamba} introduced a bidirectional scanning mechanism to model image sequences and enlarge the effective receptive field. PlainMamba~\citep{PlainMamba} further adopted a four-directional scanning strategy to enhance contextual modeling in images.

Recently, the efficiency and long-range modeling capabilities of SSMs and RNN-like architectures have been extensively explored across broader visual tasks. For instance, MambaAD~\citep{mambaAD2024} utilizes state space models to capture complex normal patterns for multi-class unsupervised anomaly detection, while recent comparative studies~\citep{mamba_rwkv2024} have investigated Mamba and RWKV architectures for high-quality and high-efficiency visual representations in vision foundation models. Motivated by these advances, several studies~\citep{UMamba,wang2024mambaunet} have integrated U-Net architectures with visual Mamba blocks and achieved promising results in medical image segmentation. The rapid evolution of SSMs has further led to advanced architectures tailored for complex medical data. For instance, PV-SSM~\citep{pvssm} explored pure visual state space models for high-dimensional medical image analysis. For volumetric data, SegMamba~\citep{segmamba2024} and SegMamba-v2~\citep{segmambav2} extended long-range sequential modeling to 3D medical image segmentation. To improve computational efficiency, UltraLight VM-UNet~\citep{ultralightvmunet} proposed a lightweight visual Mamba design for skin lesion segmentation.

However, many existing visual Mamba methods still rely on predefined scanning patterns, which may limit their ability to preserve fine-grained local structures in images with irregular boundaries. To address local structure modeling, LocalMamba~\citep{huang2024localmamba} introduced local window-based scanning to capture local dependencies while maintaining global context. To reduce the limitation of manually designed scanning patterns, DefMamba~\citep{liu2025defmamba} proposed a deformable visual state space model for adaptively capturing spatial contexts in general vision tasks. Different from these works, FU-Mamba focuses on 2D dental image segmentation and integrates content-adaptive dynamic sampling with frequency-domain enhancement. This design aims to address dental-specific challenges, including complex tooth boundaries, low-contrast regions, uneven illumination, and reflection-induced frequency imbalance, thereby improving the modeling of fine-grained two-dimensional oral structures.

\subsection{Frequency Domain Analysis}
In recent years, frequency-domain analysis has garnered considerable interest within the domain of computer vision, especially regarding its contribution to improving the robustness and generalization capabilities of deep learning models. Early studies have shown that convolutional neural networks (CNNs) exhibit an implicit bias toward high-frequency components, making them vulnerable to adversarial perturbations. For example, Caro et al.~\citep{caro2020local} demonstrated that local convolutions inherently favor high-frequency features, leading to vulnerabilities in adversarial settings. 
To address these challenges, researchers have explored approaches for balancing frequency components within neural networks. Zhang et al.~\citep{zhang2024mitigating} introduced the high-frequency feature disentanglement and recalibration (HFDR) module, which strategically isolates and recalibrates frequency-specific features to mitigate low-frequency bias and enhance adversarial robustness. In addition, wavelet-based approaches have been proposed to capture multi-scale frequency information. For instance, WTConv~\citep{WTConv} utilizes wavelet transforms to expand the receptive field without excessive parameterization, thereby enhancing the modeling of low-frequency features. Meanwhile, SPAM~\citep{SPAM} further designed frequency-aware convolutional mixers that dynamically adjust feature fusion based on the spectral distribution.
Despite these advancements, existing approaches still face limitations in effectively balancing frequency components, especially under complex imaging conditions such as dental images, which are prone to noise and uneven illumination. These challenges highlight the need for more adaptive approaches that can dynamically respond to varying frequency characteristics.

\begin{figure*}[htbp]
	\centering
	{\includegraphics[width=0.95 \linewidth]{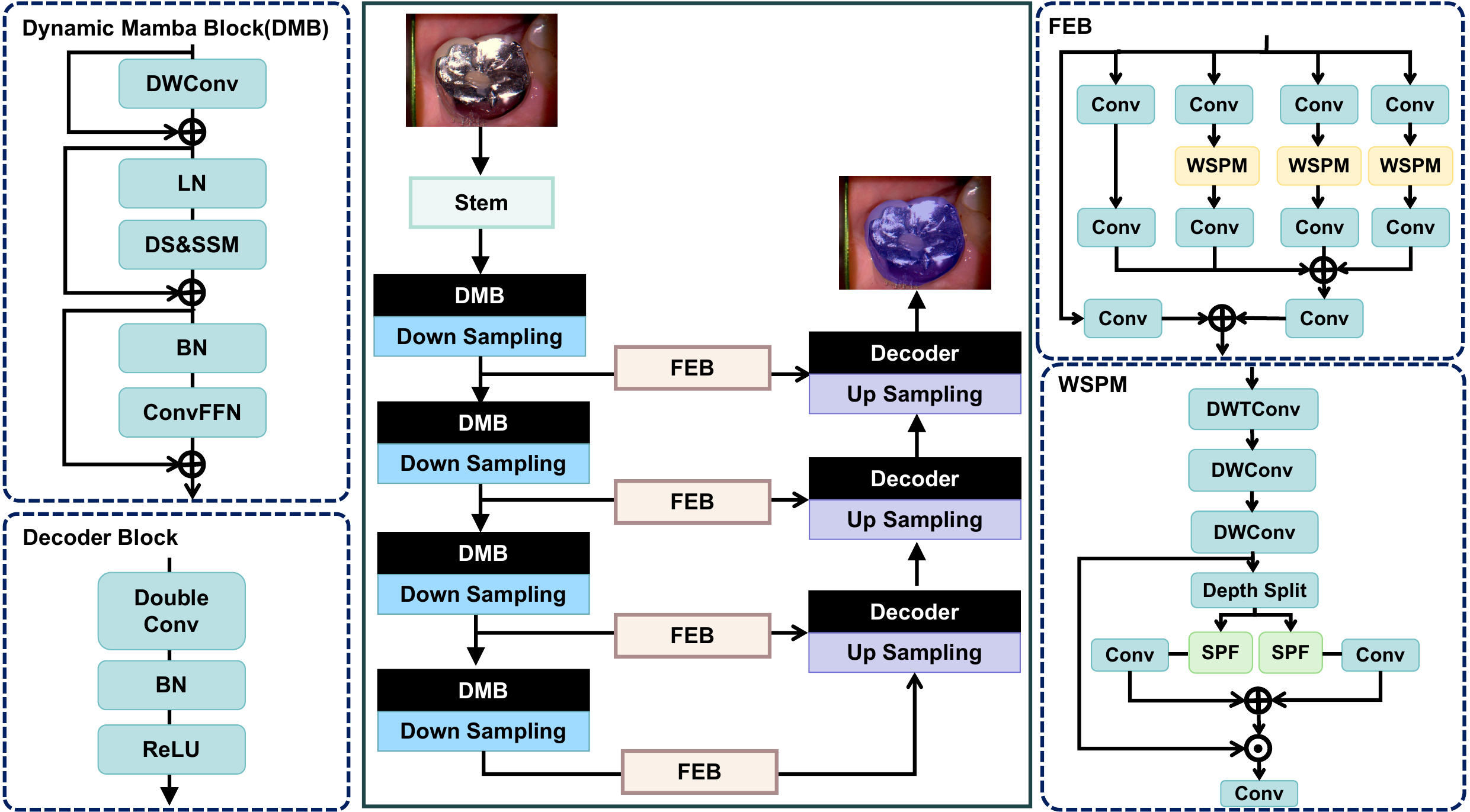}}\\ 
	\caption{\textbf{The comprehensive framework of the proposed approach.}
		The proposed approach follows the UNet architecture and comprises multiple encoders, decoders, and skip connections. In particular, the skip connections are constructed using frequency domain enhancement blocks. Meanwhile, encoders are built upon DMBs, ensuring effective feature extraction and reconstruction. In the figure, FEB refers to the frequency domain enhancement block, WSPM denotes a wavelet-guided spectral pooling module, and SPF represents the spectrum pooling filter. The blue area represents the mask generated after image segmentation. }\label{method}
\end{figure*}

\section{Our Approach} \label{ourApproach}
We design a dental image segmentation approach based on the UNet architecture, which takes dental images as input and outputs corresponding segmentation masks. The fundamental principles of state space models are introduced in Section \ref{A}. Specifically, the dynamic scanning module is employed for feature extraction (see Section \ref{B}), while the FEB is used to effectively balance high- and low-frequency features (see Section \ref{C}). This overall design facilitates the generation of high-quality segmentation masks. The detailed architecture is illustrated in Fig. \ref{method}.

\subsection{Preliminaries: State Space Model}\label{A}
SSMs were originally developed to analyze dynamic systems and time-series data, making it inherently unsuitable for processing two-dimensional images that lack causal relationships. To address this limitation, images are often converted into sequences, where scanning mechanisms help capture spatial features and transform them into visual representations. In SSMs, the temporal progression of the system's states is represented by a hidden latent state denoted as $\mathbf{h}(t) \in \mathbb{R}^{N}$. This latent state facilitates the mapping of an input sequence $\mathbf{x}(t) \in \mathbb{R}$ to an output sequence $\mathbf{y}(t) \in \mathbb{R}$. The dynamics of state transitions and observations are described by the subsequent linear differential equations:
\begin{eqnarray}
	\mathbf{h^{\prime}(t)=\alpha h(t)+\beta x(t),}
\end{eqnarray}
\begin{eqnarray}
	\mathbf{y(t)=\theta h(t)+\delta x(t).}
\end{eqnarray}
where the parameter matrices $\boldsymbol{\alpha} \in \mathbb{C}^{N \times N}$, $\boldsymbol{\beta} \in \mathbb{C}^{N}$, $\boldsymbol{\theta} \in \mathbb{C}^{N}$, and $\boldsymbol{\delta} \in \mathbb{C}^{1}$  define the dependencies between system states, inputs, and outputs.

To enhance computational efficiency and adapt SSM for sequence modeling, the Mamba framework introduces a discretization step by incorporating a time-step parameter 
$\Delta t$. 
This transformation facilitates the conversion of continuous system parameters, denoted as $\boldsymbol{\alpha}$ and $\boldsymbol{\beta}$, into their discrete equivalents, represented as $\overline{\boldsymbol{\alpha}}$ and $\overline{\boldsymbol{\beta}}$. A commonly utilized technique for executing this conversion is the zero-order hold (ZOH) approach, which is mathematically characterized as follows:
\begin{eqnarray}
	\mathbf{\overline{\alpha}=\exp (\Delta \mathbf{\alpha}),}
\end{eqnarray}
\begin{eqnarray}
	\mathbf{\overline{\beta}=(\Delta \mathbf{\alpha})^{-1}(\exp (\Delta \mathbf{\alpha})-\mathbf{I}) \cdot \Delta \mathbf{\beta}.}
\end{eqnarray}

With this discretization, the system dynamics are reformulated as follows:
\begin{eqnarray}
	\mathbf{h_{t}=\overline{\alpha} h_{t-1}+\overline{\mathbf{\beta}}x_{t},}
\end{eqnarray}
\begin{eqnarray}
	\mathbf{y_{t}=\mathbf{\theta}h_{t}+\mathbf{\delta}x_{t}.}
\end{eqnarray}

Unlike traditional SSMs, Mamba adapts parameters $\boldsymbol{\beta}$ and $\boldsymbol{\theta}$ dynamically with input.
Consequently, this modification ensures that Mamba maintains linear computational complexity during forward propagation, making it more efficient in sequence modeling tasks.

\begin{figure*}[t]
	\centering
	\includegraphics[width=0.95\linewidth]{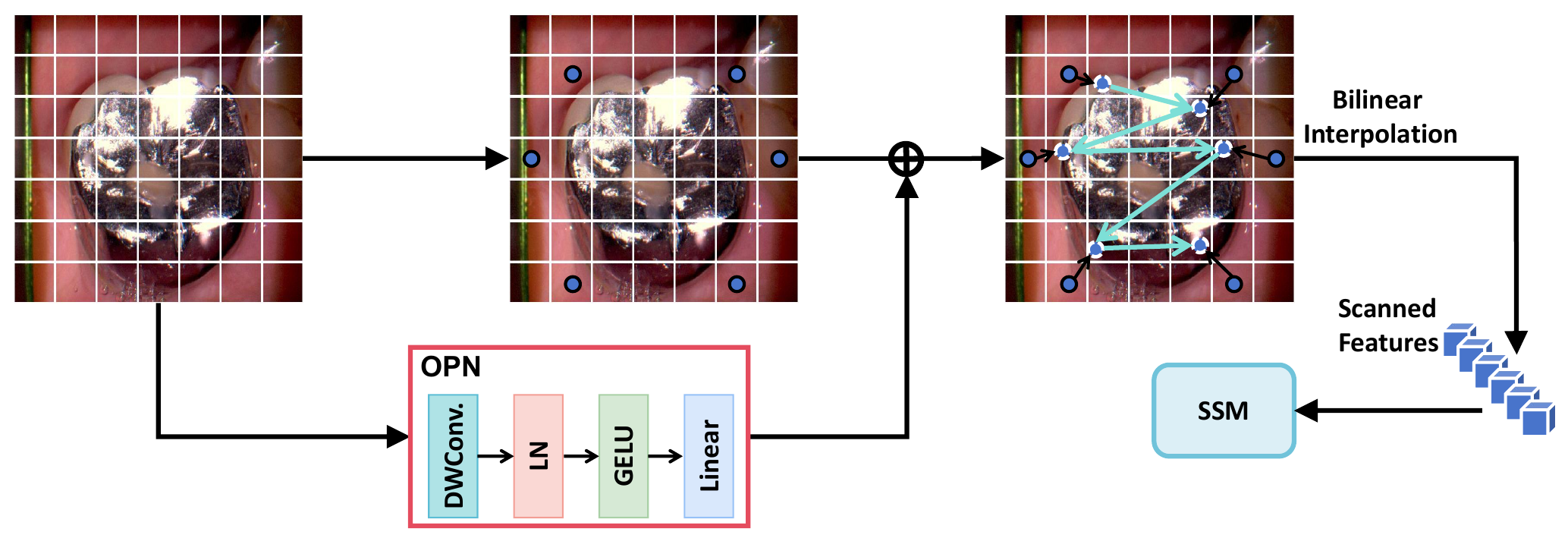}\\
	\caption{\textbf{Illustration of the DMB, where only six reference points are displayed for clarity.} Each starting reference point corresponds to a patch's origin, with its offset calculated via the OPN. Utilizing the predicted two-dimensional coordinates, bilinear interpolation is applied to extract features from significant areas, which are subsequently input into the SSM. To demonstrate our scanning approach, the figure displays six sample points, though the full process includes additional points.
		\label{figScan}}
\end{figure*}
\subsection{Dynamic Mamba Block (DMB)}\label{B}
In visual data processing, images are typically represented in a two-dimensional structure. However, manually designed scanning patterns tailored for two-dimensional images often overlook the continuous correlation between image patches. Although continuous scanning maintains consistency between neighboring patches, it often sacrifices fine-grained spatial details. On the other hand, localized scanning techniques preserve spatial relationships but struggle to establish connections across distant regions. These approaches depend on rigid, predetermined scanning patterns that fail to adjust dynamically to an image's unique features.

To address this constraint, as illustrated in Fig.~\ref{figScan}, we introduce a dynamic scanning technique that adjusts the search areas based on the input image's features. This approach enables the model to prioritize key foreground details while disregarding irrelevant background noise~\citep{ong2025fixing}, resulting in a smarter, more targeted scanning process that adapts to the visual context.
Moreover, the proposed DMB is integrated across multiple feature scales, progressively reducing spatial resolution while increasing feature dimensionality at each stage. Within each DMB, depthwise convolutions (DWConv) and convolutional feed-forward networks (ConvFFN) are employed to enhance local feature extraction, thereby facilitating more precise spatial modeling and improving the overall representational capacity of the network.

\textbf{Dynamic scanning (DS).} DS is a mechanism designed to model the relationships 
between image patches by dynamically adjusting sampling locations based on key regions 
in the feature map. The core idea is to leverage an offset prediction network (OPN) to 
learn content-aware sampling positions, extract corresponding features from the feature 
map using bilinear interpolation, and then feed them into the SSM for feature aggregation. 
Since the objective of the OPN is to regress smooth and spatially coherent offsets, it is intentionally designed as a lightweight module to ensure efficiency and stable optimization.
Specifically, the OPN adopts a lightweight architecture consisting of a depthwise convolution layer, followed by Layer Normalization (LN), a GELU activation function, and a linear projection layer for offset regression. The depthwise convolution employs a $3\times3$ kernel to enhance local spatial sensitivity while maintaining low computational complexity. The predicted offsets are further constrained within the normalized coordinate range using a \texttt{tanh} operation to stabilize dynamic sampling and avoid abrupt spatial shifts during training. To prevent invalid sampling positions, all predicted coordinates are clipped to the valid feature map range before bilinear interpolation. Coordinates exceeding the boundary are projected back to the nearest valid location, ensuring stable feature extraction near tooth boundaries and low-contrast regions.

Specifically, DS first employs OPN to predict the coordinate offsets for each image patch relative to its original position:
\begin{eqnarray}
	\Delta r = OPN([x_1,x_2,...,x_n]).
\end{eqnarray}

The offsets identify which areas warrant greater focus, directing where features should be sampled. In mathematical terms, OPN processes the feature map and generates coordinate offsets ($\Delta$ r) for every patch. These offsets are combined with the initial coordinates (p) to compute the adjusted sampling positions (p$^{\prime}$), as shown below:  
\begin{eqnarray}  
	p^{\prime}[h,w,:] = p[h,w,:] + \Delta r[h,w,:]. 
\end{eqnarray}  
Here, $h$ and $w$ denote the patch’s positional indices along the height and width axes. To maintain consistency across varying feature map resolutions, these coordinates are normalized to fall within the range [-1, 1].

After identifying the updated sampling points $p^{\prime}$, the DS module employs bilinear interpolation to fetch relevant features from the feature map. For a predicted sampling coordinate $p^{\prime}_{h,w}=(a_x,a_y)$, the sampled feature is computed as:
\begin{equation}
	X^{\prime}[h,w] = \phi(p^{\prime}_{h,w}, X),
	\label{equ9}
\end{equation}
where $\phi(\cdot)$ denotes the bilinear interpolation function. Specifically, it is formulated as:
\begin{equation}
	\phi(p^{\prime}_{h,w}, X) =
	\sum_{(r_x,r_y)\in \mathcal{N}(p^{\prime}_{h,w})}
	g(a_x,r_x,a_y,r_y) X[r_y,r_x,:],
	\label{equ10}
\end{equation}
where $\mathcal{N}(p^{\prime}_{h,w})$ denotes the four nearest integer grid points around $p^{\prime}_{h,w}$. The interpolation weight is defined as:
\begin{equation}
	g(a_x,r_x,a_y,r_y) =
	\max(0, 1 - |a_x - r_x|) \times \max(0, 1 - |a_y - r_y|).
	\label{equ11}
\end{equation}

Here, $a_x$ and $a_y$ represent the horizontal and vertical coordinates of the predicted sampling position, while $r_x$ and $r_y$ denote the indices of neighboring feature map locations. The function $g(a_x,r_x,a_y,r_y)$ computes the bilinear interpolation weight according to the spatial distance between the predicted coordinate and its neighboring grid points.

\begin{algorithm}[htbp]
\caption{Dynamic Scanning}
\label{alg:dynamic_scanning}
\begin{algorithmic}[1]
\REQUIRE Input feature map $X \in \mathbb{R}^{H \times W \times C}$
\ENSURE Reordered feature sequence $X' \in \mathbb{R}^{N \times C}$

\STATE Predict offsets:
$
\Delta r \gets \mathrm{OPN}(X)
$
\STATE Compute updated coordinates:
$
p' \gets p + \Delta r
$

\STATE Bilinear feature sampling:
$
f[h,w] \gets \phi(p'[h,w], X)
$

\STATE Generate semantic-aware ordering:
$
\mathcal{I} \gets \mathrm{sort}(p')
$

\FOR{$k=1$ to $N$}
    \STATE $(h_k,w_k)\gets\mathcal{I}_k$
    \STATE $X'[k]\gets f[h_k,w_k]$
\ENDFOR

\RETURN $X'$
\end{algorithmic}
\end{algorithm}

After feature sampling, the collected features are reordered according to the spatial proximity of the updated coordinates $p'$. Specifically, neighboring sampled positions in the dynamic coordinate space are arranged adjacently in the generated sequence, allowing the SSM to better preserve local structural continuity during sequential modeling.
After collecting the sampled features, DS reorganizes them according to their updated spatial coordinates before feeding them into the SSM for sequential modeling. To adaptively extract informative regions, we employ a dynamic scanning mechanism, whose complete workflow is outlined in Algorithm \ref{alg:dynamic_scanning}. Notably, the learned sampling positions in DS provide additional relative position information, which helps the SSM better capture local feature relationships and improve overall feature aggregation capability.
In summary, DS dynamically adjusts sampling locations, allowing the feature extraction process to flexibly focus on key regions while using bilinear interpolation to achieve smooth information aggregation. 

\subsection{Frequency Domain Enhancement Block}\label{C}
The segmentation of dental images poses distinct challenges attributable to the intrinsic complexity of their frequency composition. On one hand, high-frequency artifacts, including scanner-induced noise and sharp discontinuities along tooth boundaries, can interfere with accurate localization. On the other hand, low-frequency structures like smooth gingival contours and the general geometry of teeth encode critical global information. Traditional approaches often fail to capture and reconcile these contrasting frequency components effectively, especially under conditions of poor contrast between teeth and soft tissue, individual anatomical variability, or inconsistent lighting. These limitations lead to degraded segmentation quality and poor model robustness across diverse cases.

In order to tackle these challenges, we propose a frequency-aware enhancement mechanism that emphasizes perceptually important structures while suppressing irrelevant high-frequency disturbances. Motivated by the contrast sensitivity profile of the human visual system, which is most responsive to mid-frequency signals, we introduce the WSPM. This module is intended to enhance low-frequency components while simultaneously integrating them with high-frequency details in a balanced manner, thus directing the spectral distribution towards mid-range frequencies.

To ensure adaptability across multiple spatial scales, the WSPM employs a multi-branch design with variable convolutional depths and kernel shapes. The low-frequency region is defined as a centered circular area in the frequency spectrum. To ensure adaptability across different feature scales, its radius is dynamically determined according to the spatial resolution of the input feature map. Specifically, the radius is set to one-eighth of the minimum feature map dimension at each encoder stage:
\begin{equation}
r = \frac{\min(H,W)}{8}.
\end{equation}
This adaptive setting enables consistent low-frequency coverage across different feature scales while avoiding excessive suppression of mid-frequency components.

Directional sensitivity is strengthened using separable convolutions with kernels sized 1×n and n×1. Simultaneously, dilated convolutions (rates: 1, 3, 5, 7) widen the receptive field. Appropriate padding ensures alignment across branches, facilitating smooth concatenation during feature aggregation. Features extracted from different branches are concatenated along the channel dimension and subsequently fused using a $1\times1$ convolution layer for adaptive channel aggregation. This adaptive fusion strategy allows the network to learn branch-wise feature importance automatically rather than relying on uniform averaging. The combination of asymmetric convolutions and different dilation rates enables the network to capture directional structures and medium-frequency contextual patterns at multiple receptive fields, thereby enhancing structural continuity while suppressing excessive high-frequency noise.

The key concept of deep wavelet convolution (DWT-Conv) combines wavelet transforms with convolutional operations to efficiently capture and refine multi-scale image features.
Specifically, DWT-Conv first applies wavelet decomposition to the input image, typically using the Haar wavelet transform. This process separates the image into low-frequency (LL) and high-frequency (LH, HL, HH) components. In our implementation, a single-level Haar wavelet decomposition is employed at each encoder stage. Fixed Haar wavelet kernels are adopted instead of learnable wavelet bases to maintain stable frequency decomposition and avoid introducing additional trainable parameters. This design also improves computational efficiency while preserving high-frequency edge structures commonly observed in dental images. This process can be represented as:
\begin{equation}
	[X_{LL}, X_{LH}, X_{HL}, X_{HH}] = \text{DWT}(X). 
\end{equation}
where \( X \in \mathbb{R}^{H \times W \times C} \) represents the input feature map. The DWT decomposes it into four subbands (\(X_{LL}, X_{LH}, X_{HL}, X_{HH}\)), each with reduced spatial dimensions and preserved channels \( \mathbb{R}^{\frac{H}{2} \times \frac{W}{2} \times C} \). Subsequently, a \( 1\times1 \) convolution operation is employed on the concatenated subbands along the channel dimension (yielding an intermediate tensor of \( \mathbb{R}^{\frac{H}{2} \times \frac{W}{2} \times 4C} \)). The convolution outputs a refined tensor of the same size \( 4C \), which is then split back into four enhanced subbands to augment low-frequency features while efficiently modeling cross-band relationships:
\begin{equation}
	[X_{LL}', X_{LH}', X_{HL}', X_{HH}'] = \text{Split}(\text{Conv}_{1\times1}(\text{Concat}(X_{LL}, X_{LH}, X_{HL}, X_{HH}))).
\end{equation}

The final step involves applying an inverse wavelet transform to combine the refined subbands, producing the enhanced feature map as follows:
\begin{equation}
	Y = \text{IWT}([X_{LL}', X_{LH}', X_{HL}', X_{HH}']).
\end{equation}

To alleviate potential information loss during inverse reconstruction, a residual connection is introduced between the input feature map and the reconstructed output. In addition, Layer Normalization is applied after subband fusion to stabilize the feature distribution under varying illumination and noisy conditions.

This operation not only enhances low-frequency information but also preserves the spatial resolution of the image to some extent, providing a richer information foundation for subsequent feature extraction and segmentation tasks. It is particularly useful for handling the complex relationships between structures such as teeth and gums in dental images.

The spectrum pooling filter (SPF) serves as a key mechanism for harmonizing high- and low-frequency data, enhancing feature representation through balanced spectral processing. By converting spatial feature maps into the frequency domain via a 2D discrete Fourier transform (DFT), the SPF dynamically regulates various frequency components. This transformation enables precise modulation of spectral information, ensuring optimal feature extraction across different frequency bands. This procedure can be illustrated as follows:
\begin{equation}
	Z = F(z),
\end{equation}
where \( z \) represents the feature map after deep convolution, and \( F \) denotes the 2D DFT operation. Then, the spectrum is centralized by focusing the low-frequency components in the center and removing the remaining parts outside the spectrum. This procedure can be illustrated as follows:
\begin{equation}
	S_{lf}(u,v) = 
	\begin{cases}
		G(Z)(u,v), & \text{if } (u,v) \in A_{lf}, \\
		0, & \text{otherwise}.
	\end{cases}
\end{equation}
\begin{equation}
	S_{hf} = G(Z) - S_{lf}.
\end{equation}
Here, \( G(\cdot) \) is the Fourier transform centralization function, and \( Al_f \) defines the low-frequency region centered at the origin. The centralization operation only reorders the spatial arrangement of frequency components through FFT shift without altering their original magnitude or phase values, thereby preserving structural boundary consistency during inverse transformation. Next, the processed frequency features are mapped back to the spatial domain through inverse transformations:
\begin{equation}
	f_{lp}(Z) = F^{-1}(G^{-1}(S_{lf})),
\end{equation}
\begin{equation}
	f_{hp}(Z) = F^{-1}(G^{-1}(S_{hf})).
\end{equation}

Finally, the features from different frequency bands are fused to generate the spectrum-pooled feature map:
\begin{equation}
	\tilde{Z} = \lambda f_{lp}(Z) + (1 - \lambda) f_{hp}(Z),
\end{equation}
where \( \lambda \) is the balancing parameter that controls the proportion of high- and low-frequency components. 







This approach allows SPF to minimize disruptions caused by high-frequency noise while maintaining critical low-frequency structural details. As a result, it significantly improves the precision and reliability of dental image segmentation, especially when dealing with complex interfaces between teeth and surrounding soft tissues.

\section{Results}  \label{results}
We assess the efficacy of our methodology and conduct a comparative analysis with alternative methodologies utilizing two distinct datasets.

\subsection{Datasets and Evaluation Criteria}
\subsubsection{Datasets}
In order to assess the efficacy of our proposed semantic segmentation approach, we performed evaluations utilizing two publicly accessible dental datasets: the DSD~\citep{TIAN2020107158} and the Oralvision~\citep{oralvision} dataset.

\textbf{ DSD} used in this study is derived from the dental segmentation data described in Tian et al.~\citep{TIAN2020107158}. Specifically, we extracted 2D dental image frames and their corresponding pixel-wise segmentation masks from the original dental video data for 2D dental image segmentation experiments. The DSD subset follows a multi-class semantic segmentation setting, where each pixel is annotated into one of several oral structure categories, including teeth, surrounding soft tissues, and background, rather than a binary tooth-background label. The resulting DSD subset consists of 4900 training images, 151 validation images, and 151 test images. The training, validation, and test sets were constructed without image overlap, and all splits were mutually exclusive to prevent potential data leakage. The detailed data augmentation strategy is described in Section 4.2.

\textbf{The Oralvision dataset} comprises intraoral images captured at a resolution of 640×480 using dental scanning equipment in clinical settings. These images were collected from randomly selected patients and manually annotated by six university researchers utilizing LabelM. The dataset categorizes oral structures into different tissue types, including gums, lips, teeth, tongue, and cheeks. Furthermore, it incorporates real-world challenges by retaining visual noise elements such as food particles, dental calculus, implant attachments, and saliva. The dataset is comprised of 4000 images designated for training purposes, 120 images allocated for validation, and an additional 120 images set aside for testing. The training, validation, and test sets were constructed without image overlap, and all splits were mutually exclusive to prevent potential data leakage. All OralVision images were resized to a unified input resolution during preprocessing while preserving challenging artifacts such as saliva, dental calculus, and illumination variations for robustness evaluation. 

\subsubsection{Evaluation Criteria}
To assess segmentation quality, we adopt mean Intersection over Union (mIoU) together with boundary mean IoU (mBIoU). These metrics respectively evaluate overall spatial overlap and edge delineation precision across all semantic categories. Per-class IoU scores are first computed and subsequently averaged to yield mIoU and mBIoU, thereby reflecting model behavior across the entire tissue taxonomy.

Boundary IoU is determined according to standard practice: for each category, its contour is extracted and dilated by a tolerance of $t=3$ pixels to form a narrow band around the boundary. Denoting the dilated boundary pixel sets of the ground-truth and predicted masks as $B_{gt}$ and $B_{pred}$, the boundary IoU is defined as
\begin{equation}
\mathrm{IoU}_{\text{boundary}} = \frac{|B_{gt}\cap B_{pred}|}{|B_{gt}\cup B_{pred}|}.
\end{equation}

We report parameter size, computational complexity, and inference latency as efficiency indicators. FLOPs were estimated at the same input resolution for all methods. For SAM-based methods, one bounding-box prompt was used during evaluation. Inference latency was measured under single-sample inference with a batch size of 1.



\subsection{Implementation Details}

All experiments were conducted on a high-performance workstation equipped with an Intel Core i9-13900KF CPU, 32 GB RAM, and four NVIDIA RTX 4090D GPUs. For the efficiency benchmarks, all methods were evaluated on a single NVIDIA RTX 4090D GPU with batch size 1 and an input resolution of 640 × 480. To ensure fair and reproducible comparison, no multi-GPU inference was used during latency measurement, and all compared methods were benchmarked under the same hardware and input settings. All methods were implemented under the PyTorch framework to ensure a unified experimental environment and fair comparison. All experiments were repeated three times with different random seeds, and the reported results correspond to the mean and standard deviation. For fair comparison, all competing methods were trained and evaluated under identical data splits and random seed settings. Official implementations were adopted for all compared methods whenever publicly available, with only necessary modifications to accommodate unified dental image input resolutions and output categories.

For training efficiency, the batch size was set to 32 for the DSD dataset and 24 for the OralVision dataset, respectively, according to GPU memory consumption and convergence stability under different input resolutions. All models were trained for 45 epochs using the AdamW optimizer with an initial learning rate of 0.01 and a weight decay of $3\times10^{-5}$. A cosine annealing learning rate schedule was adopted throughout training to gradually reduce the learning rate and improve optimization stability.

The overall optimization objective consists of a hybrid segmentation loss and an offset regularization term. Specifically, the segmentation loss combines Dice loss and cross-entropy loss to jointly optimize region overlap and pixel-wise classification accuracy:
\begin{equation}
\mathcal{L}_{seg} = \mathcal{L}_{Dice} + \mathcal{L}_{CE}.
\end{equation}
Dice loss was introduced to alleviate the class imbalance commonly observed in multi-class dental image segmentation.

To stabilize dynamic sampling, an additional offset regularization loss was introduced:
\begin{equation}
\mathcal{L}_{reg} = \frac{1}{N}\sum_{i=1}^{N}\|\Delta r_i\|_2^2,
\label{equ24}
\end{equation}
where $\Delta r_i$ denotes the predicted offset at the $i$-th spatial location. The final optimization objective is defined as: \begin{equation} \mathcal{L} = \mathcal{L}_{seg} + \lambda_{\mathrm{reg}} \mathcal{L}_{reg},\end{equation}
where $\lambda_{\mathrm{reg}}$ was empirically set to 0.01.

An early stopping strategy was adopted to mitigate overfitting, where training was terminated if the validation performance did not improve for 8 consecutive epochs. The minimum improvement threshold for validation mIoU was set to 0.001. In addition, gradient clipping with a threshold of 1.0 was applied during optimization to further improve training stability.

To improve robustness against intra-oral viewpoint variations and illumination changes, data augmentation strategies including random horizontal flipping, random rotation, random scaling, and brightness adjustment were applied during training. The probability of applying each augmentation operation was set to 0.5, while the brightness adjustment factor was randomly sampled within the range of [0.8, 1.2]. Elastic deformation and CutMix augmentation were not adopted because they may introduce unrealistic anatomical distortions in dental structures.

For the proposed dynamic scanning module, the OPN adopts a lightweight architecture consisting of a $3\times3$ depthwise convolution layer, followed by LN, GELU activation, and a linear projection layer for offset regression. To stabilize dynamic sampling, the predicted offsets were normalized using a $\tanh$ activation function and constrained within the range $[-0.25,0.25]$ relative to the normalized coordinate grid. In addition, the offset regularization term in Eq. (\ref{equ24}) further constrains excessive spatial displacement and improves the smoothness of neighboring sampling locations during optimization. In addition, all predicted sampling coordinates were clipped to the valid feature map range before bilinear interpolation to avoid invalid boundary sampling.

For the DMB design, the channel dimensions across the four encoder stages were set to $\{64,128,256,512\}$, respectively. The ConvFFN expansion ratio was set to 4, while all depthwise convolutions employed a kernel size of $3\times3$.

For the frequency enhancement module, the WSPM adopted a cascaded wavelet convolution structure with a $1\times1$ convolution kernel and stride 1. Two parallel SPF branches were introduced, with frequency balancing weights empirically set to $\lambda_1=0.7$ to balance low-frequency structural information and high-frequency detail preservation. All DWT and DFT operations were performed on downsampled deep feature maps to reduce computational overhead while preserving frequency-aware representation capability.

\subsection{Ablation Study}
In order to assess the efficacy of this module, we undertook comprehensive experiments designed to examine its influence on the overall performance of the model. Specifically, we executed ablation studies utilizing the dental segmentation dataset and Oralvision dataset to evaluate the module's role in enhancing segmentation accuracy.

\textbf{Backbone.} To evaluate the effectiveness of our proposed methodology, we employ U-Mamba as the foundational architecture for our ablation studies. U-Mamba features a symmetric encoder-decoder structure that combines the strong feature extraction capabilities of UNet with the extensive context modeling benefits of the Mamba framework. This architectural design has proven to be effective in medical image segmentation tasks.

\begin{table}[htbp]
\centering
\small
\caption{\textbf{Ablation study on the scanning design.} CS refers to the continuous scanning strategy, LS denotes the local scanning strategy, and DMB denotes the proposed Dynamic Mamba Block. Results are reported as mean $\pm$ standard deviation over three independent runs.}
\begin{tabular}{lcc}
\toprule
\textbf{Scanning Strategy} & \textbf{DSD (mIoU)} & \textbf{OralVision (mIoU)}\\
\midrule
CS  & 89.7 $\pm$ 0.4 & 89.1 $\pm$ 0.5 \\
LS  & 90.2 $\pm$ 0.3 & 89.8 $\pm$ 0.4 \\
DMB & \cellcolor{blue!10}\textbf{90.9 $\pm$ 0.3} & \cellcolor{blue!10}\textbf{90.4 $\pm$ 0.3} \\
\bottomrule
\end{tabular}
\label{tab:DMB}
\end{table}

\textbf{Impact of the DMB.} 
In order to assess the impact of the DMB, we conduct a comparative analysis with two alternative scanning methodologies: continuous scanning (CS) and local scanning (LS), while keeping all other components unchanged. As shown in Table \ref{tab:DMB}, DMB markedly enhances the model's capacity to discern intricate details and adjust to spatial variations present in dental images. Specifically, it achieves a 1.2\% increase in mIoU compared to CS on the DSD. This improvement is mainly attributed to DMB's capacity to dynamically adjust sampling positions based on input features, thereby enhancing structural awareness and local context modeling. As a result, DMB enhances the precision of boundary delineation and elevates segmentation performance within intricate oral environments.

\begin{figure}[htbp]
\vspace{-4mm}
  \centering
  \subfloat[Dental Segmentation Dataset]{\includegraphics[width=0.35\linewidth]{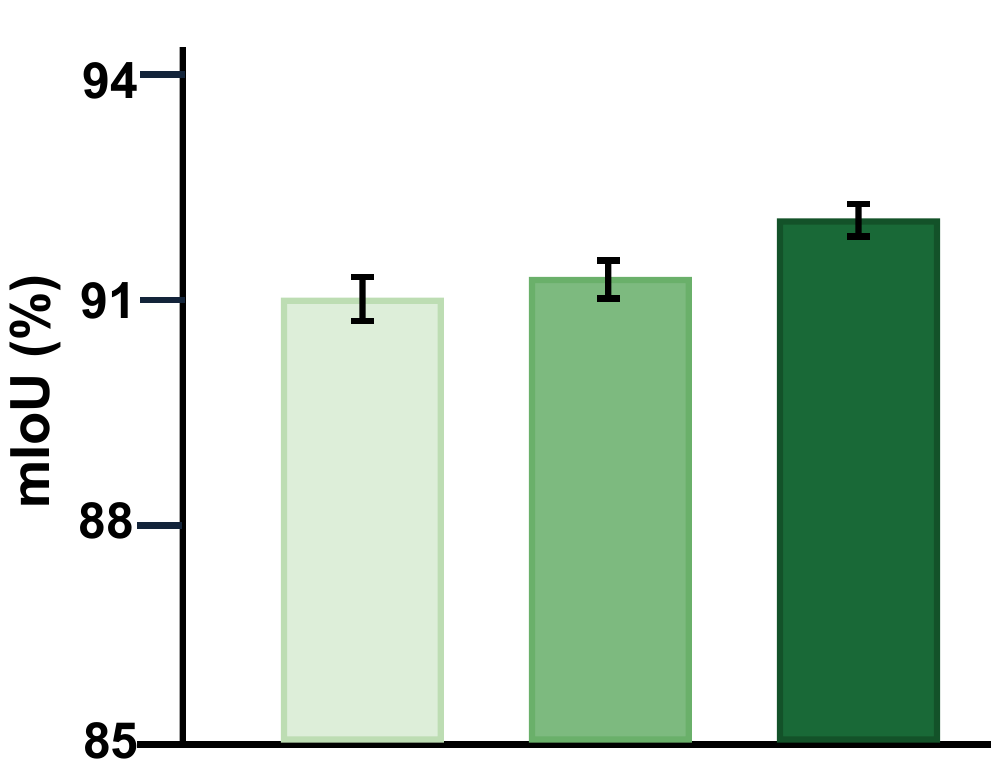}}
  \hfil
  \subfloat[OralVision Dataset]{\includegraphics[width=0.35\linewidth]{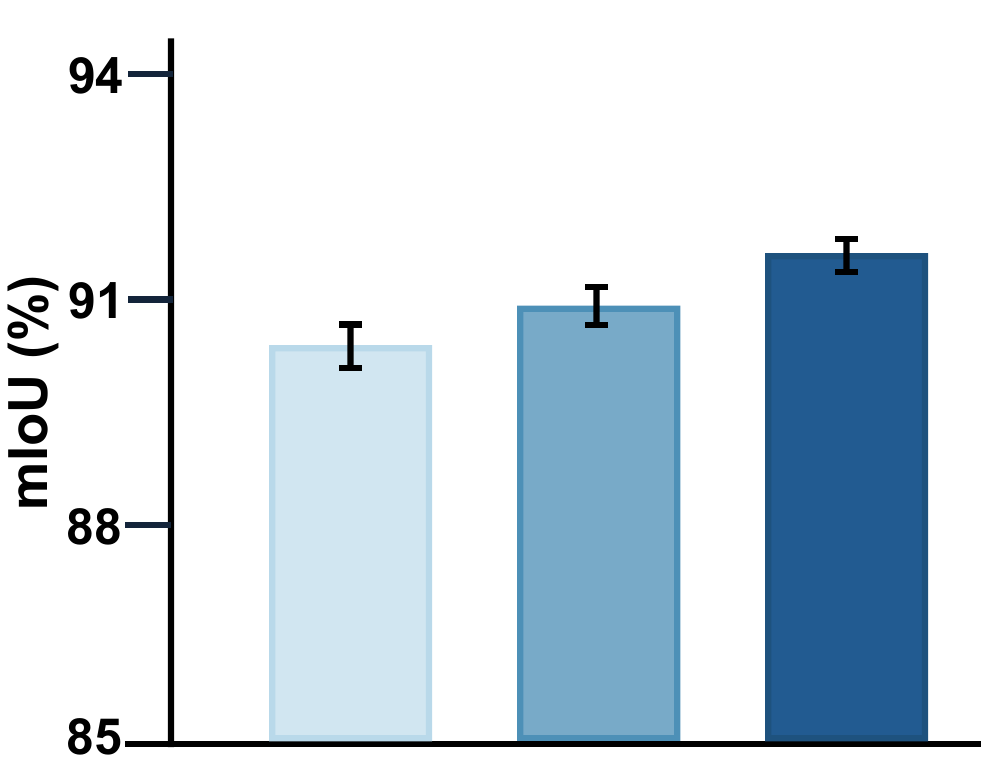}}
  \caption{\textbf{Comparison of mIoU scores on the dental segmentation dataset and the OralVision dataset with and without the FEB, as well as with the FEB-w/o WSPM variant.} Here, FEB-w/o WSPM refers to a configuration in which the WSPM component within FEB is replaced by DWT-Conv. As shown, the model incorporating the full frequency domain enhancement block achieves higher mIoU than the other two configurations.
}
  \label{feb}
  \vspace{-4mm}
\end{figure}

\begin{figure}[htbp]
    \centering
    \includegraphics[width=0.8 \linewidth]{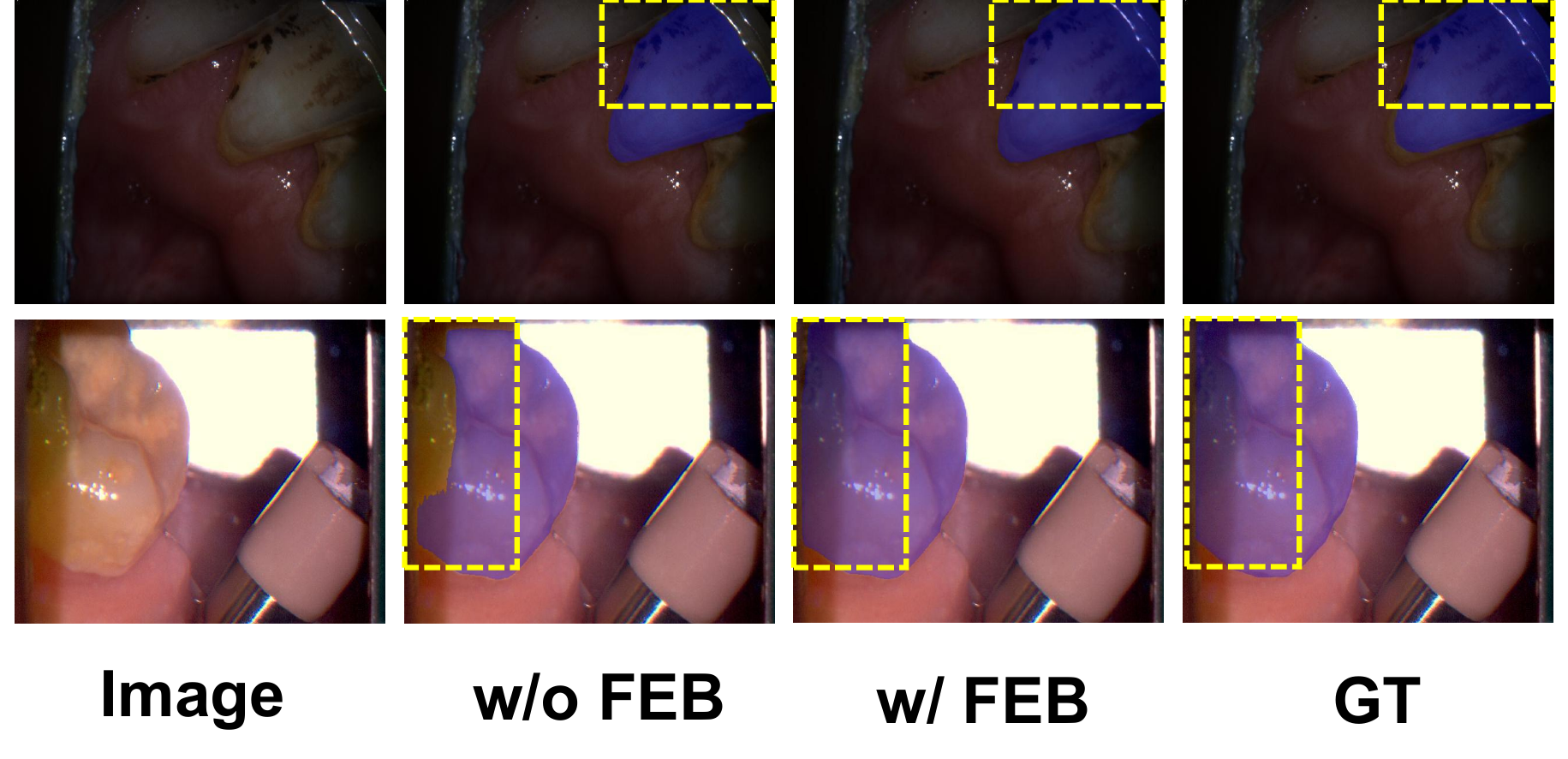}
    \caption{\textbf{Visualization comparison of the frequency domain enhancement block.} The FEB module enables the segmentation results on low-light dental images to more closely align with the ground truth. The blue regions indicate the predicted segmentation results, while the yellow dashed boxes highlight differences across various settings, emphasizing the advantages of incorporating FEB under challenging lighting conditions.}
    \label{figXr}
    \vspace{-2mm}
\end{figure}

\textbf{Impact of the Frequency Domain Enhancement Block.}
We assess the efficacy of the FU-Mamba model by comparing its performance with and without the FEB on both the dental segmentation dataset and the OralVision dataset. As shown in Fig.~\ref{feb}(a), the leftmost result corresponds to the FU-Mamba model with only skip connections, the center represents a variant named FEB-w/o WSPM in which the WSPM component within FEB is replaced by DWT-Conv, and the rightmost result shows our full version incorporating WSPM. The same comparison is presented in Fig.~\ref{feb}(b). The FU-Mamba model without FEB achieves 1.4\% lower mIoU compared to the version with the FEB. Similarly, the FEB-w/o WSPM variant also underperforms the full version that includes WSPM.
We further visualize the segmentation results with and without the FEB module. As illustrated in Fig.~\ref{figXr}, removing FEB leads to a notable degradation in segmentation quality, particularly in regions affected by noise and insufficient illumination. The observed change in performance can be ascribed to the module's capacity to equilibrate high- and low-frequency characteristics, thereby efficiently attenuating high-frequency noise while maintaining critical structural information.
 
To further investigate the effectiveness of different components within the proposed FEB, we conduct an additional ablation study on the DSD dataset, as summarized in Table~\ref{tab:feb_ablation}. Specifically, we separately evaluate the contributions of the wavelet-based DWT-Conv module and the SPF module.

Starting from the baseline model without FEB, introducing only the DWT-Conv module improves the mIoU from 90.9\% to 91.7\%, demonstrating that wavelet-guided decomposition effectively enhances multi-scale structural representation and frequency-aware feature extraction for complex dental regions. When only the SPF module is employed, the mIoU increases to 91.4\%, indicating that adaptive spectral balancing helps suppress high-frequency noise and improves semantic consistency under challenging illumination conditions.

By jointly integrating DWT-Conv and SPF, the complete FEB achieves the best performance with 92.3\% mIoU and 88.9\% mBIoU. These results demonstrate that the two components provide complementary benefits, where DWT-Conv serves as the primary contributor for frequency decomposition and structural preservation, while SPF further enhances discriminative spectral fusion and robustness against noise interference. Consequently, the combined design achieves more accurate and stable dental image segmentation.

\begin{table}[htbp]
\centering
\caption{Ablation study on the effectiveness of different frequency enhancement components in FEB.}
\label{tab:feb_ablation}
\begin{tabular}{cccc}
\toprule
SPF & DWT-Conv &  mIoU (\%) & mBIoU (\%) \\
\midrule
 &  & 90.9 $\pm$ 0.3 & 86.9 $\pm$ 0.2 \\
\checkmark &  & 91.4 $\pm$ 0.3 & 87.5 $\pm$ 0.3 \\
 &\checkmark  & 91.7 $\pm$ 0.2 & 87.9 $\pm$ 0.3 \\
\checkmark & \checkmark & \cellcolor{blue!10}\textbf{92.3 $\pm$ 0.2} & \cellcolor{blue!10}\textbf{88.9 $\pm$ 0.3} \\
\bottomrule
\end{tabular}
\end{table}

\textbf{Hyperparameter Analysis.}
To further evaluate the sensitivity of the proposed frequency enhancement mechanism, we conducted a hyperparameter analysis on the frequency balancing weight $\lambda$ in the Spectrum Pooling Filter (SPF). The parameter $\lambda$ controls the relative contribution between low-frequency structural information and high-frequency detail features during frequency-domain fusion. Specifically, a smaller $\lambda$ emphasizes high-frequency components, while a larger $\lambda$ strengthens low-frequency semantic information.

\begin{figure}[htbp]
    \centering
    \includegraphics[width=0.45\linewidth]{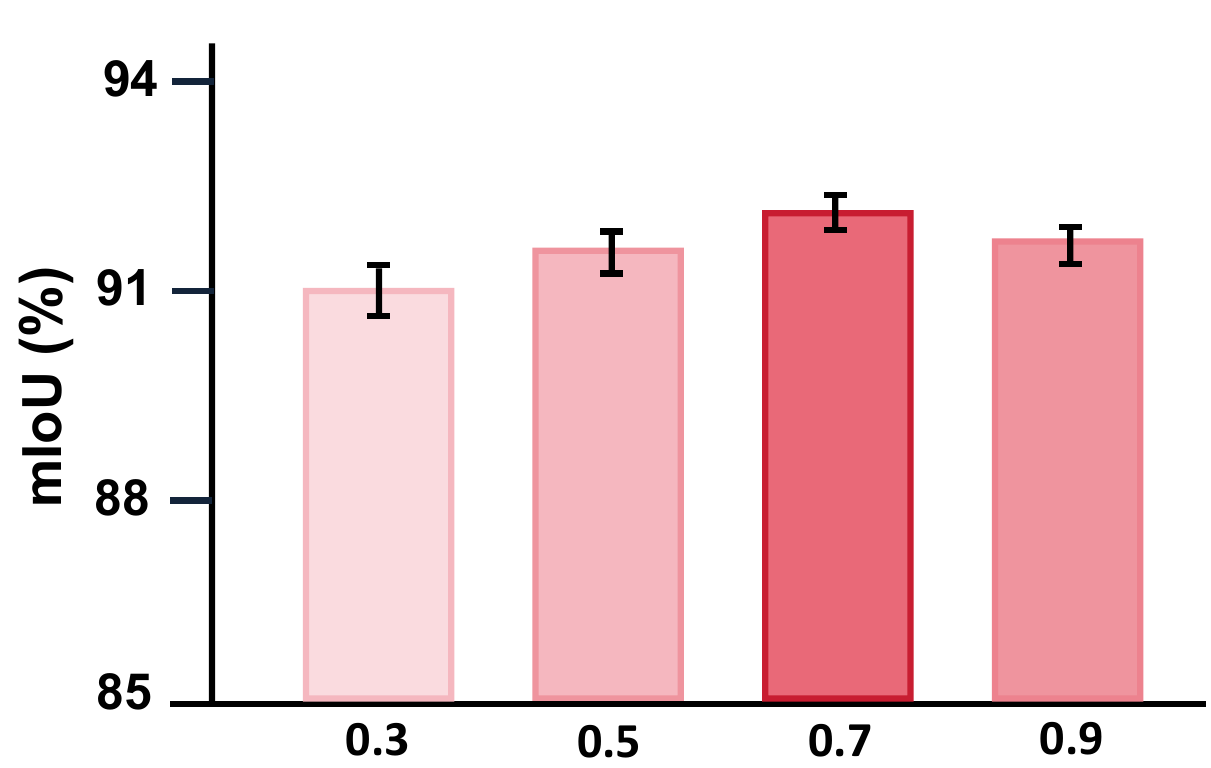}
    \caption{Effect of different frequency balancing weights $\lambda$.}
    \label{fig:lambda}
\end{figure}

Fig.~\ref{fig:lambda} reports the segmentation performance under different $\lambda$ settings on the DSD dataset. When $\lambda$ is set to 0.3, the model tends to overly focus on high-frequency details and becomes more sensitive to noise interference, resulting in relatively lower segmentation accuracy. As $\lambda$ increases, the model achieves better structural consistency and boundary representation. The best performance is obtained at $\lambda=0.7$, achieving an mIoU of 92.3\%. However, when $\lambda$ is further increased to 0.9, excessive emphasis on low-frequency information weakens fine-grained boundary modeling capability, leading to a slight performance decrease.

\textbf{Incremental Ablation and Complexity Analysis.}
To further evaluate the effectiveness of each proposed component, an incremental ablation study is conducted on the proposed FU-Mamba framework. Starting from the baseline U-Mamba architecture, the DMB and FEB are progressively introduced to analyze both segmentation performance improvement and computational overhead. The quantitative results on the DSD dataset are summarized in Table~\ref{tab:complexity}.

\begin{table}[htbp]
\centering
\small
\caption{Incremental ablation and complexity analysis on the DSD.}
\label{tab:complexity}
\begin{tabular}{cccccc}
\hline
DMB & FEB & mIoU (\%)  & mBIoU (\%) & Params (M)  & Latency (ms)  \\
\hline
 &  & 90.1 $\pm$ 0.3 & 85.8 $\pm$ 0.4 & 32.8 & 21.3 \\
\checkmark &  & 90.9 $\pm$ 0.3 & 86.7 $\pm$ 0.3 & 34.1 & 25.7 \\
\checkmark & \checkmark & \cellcolor{blue!10}\textbf{92.3 $\pm$ 0.2} & \cellcolor{blue!10}\textbf{88.9 $\pm$ 0.3} & \cellcolor{blue!10}\textbf{36.2} & \cellcolor{blue!10}\textbf{31.8} \\
\hline
\end{tabular}
\end{table}
Specifically, after introducing the DMB module, the model achieves noticeable improvement in mIoU owing to the adaptive dynamic scanning strategy, which enhances spatial continuity modeling and global context aggregation. 

When further incorporating the FEB module, the segmentation accuracy is further improved, particularly in complex boundary regions and low-contrast structures. This improvement mainly benefits from the explicit modeling of frequency-domain information through wavelet decomposition and spectrum-aware enhancement. Meanwhile, the additional computational overhead is primarily introduced by the DWT and DFT operations in the frequency-domain branch.

\subsection{Evaluation of Tooth Segmentation}
We conduct a comparative analysis of our methodology against various leading segmentation techniques, including those based on SAM~\citep{sam,sam-hq,efficientsam,groundedsam} and Mamba-like architectures~\citep{VisionMamba,vrwkv,UMamba,wang2024mambaunet}, under consistent experimental settings. We also include T-Mamba~\citep{tmamba}, a unified frequency-based Mamba framework for 2D and 3D tooth segmentation, to provide a domain-specific baseline.

\begin{table}[htbp]
	\centering
	\small
    \caption{\textbf{Experimental results on the test set of DSD and the Oralvision dataset. }Our experimental results are highlighted in purple.}
	\begin{tabular}{lcccc}
		\toprule
		\multirow{2}{*}{Approach} & \multicolumn{2}{c}{DSD} & \multicolumn{2}{c}{OralVision} \\
		\cmidrule(lr){2-3} \cmidrule(lr){4-5}
		& mIoU & mBIoU & mIoU & mBIoU \\
		\midrule
		SAM~\citep{sam}   & 88.7 ± 0.4 & 85.3 ± 0.5 & 88.1 ± 0.3 & 85.0 ± 0.4 \\
		HQ-SAM~\citep{sam-hq}      & 91.2 ± 0.3 & 88.1 ± 0.4 & 90.3 ± 0.3 & 87.2 ± 0.3 \\
		EfficientSAM~\citep{efficientsam} & 87.2 ± 0.5 & 84.4 ± 0.5 & 86.9 ± 0.4 & 83.8 ± 0.3 \\
		GroundedSAM~\citep{groundedsam} & 88.6 ± 0.4 & 85.5 ± 0.3 & 87.6 ± 0.3 & 84.7 ± 0.3 \\
        \midrule
        nnU-Net~\cite{nnunet} & 89.7 ± 0.3 & 86.1 ± 0.3 & 89.3 ± 0.3 & 85.6 ± 0.4 \\
        U-KAN~\cite{ukan} & 89.1 ± 0.4 & 85.4 ± 0.3 & 88.7 ± 0.4 & 85.0 ± 0.4 \\
        MedNeXt~\cite{mednext} & 90.2 ± 0.3 & 86.6 ± 0.4 & 90.0 ± 0.3 & 86.1 ± 0.3 \\
        \midrule
        T-Mamba~\cite{tmamba} & 90.4 ± 0.3 & 86.8 ± 0.3 & 89.8 ± 0.3 & 86.0 ± 0.4 \\
		Vim~\citep{VisionMamba} & 90.6 ± 0.3 & 86.4 ± 0.4 & 89.9 ± 0.4 & 86.3 ± 0.3 \\
		VRWKV~\citep{vrwkv} & 89.8 ± 0.4 & 85.7 ± 0.3 & 89.2 ± 0.3 & 84.8 ± 0.3 \\
		U-Mamba~\citep{UMamba} & 90.1 ± 0.3 & 85.8 ± 0.4 & 90.0 ± 0.3 & 85.2 ± 0.3 \\
        Mamba-UNet~\citep{wang2024mambaunet}& 90.3 ± 0.2 & 86.0 ± 0.3  & 89.7 ± 0.3 & 84.9 ± 0.4 \\
		\midrule
		 \textbf{Ours(FU-Mamba)}  & \cellcolor{blue!10}\textbf{92.3 ± 0.2} & \cellcolor{blue!10}\textbf{88.9 ± 0.3} & \cellcolor{blue!10}\textbf{91.6 ± 0.3} & \cellcolor{blue!10}\textbf{88.0 ± 0.4} \\
		\bottomrule
	\end{tabular}
	\label{tabSegmentation}
\end{table}

In a fair comparison under identical experimental conditions, our algorithm demonstrated superior performance on the DSD and Oralvision dataset compared to the aforementioned approaches in Table \ref{tabSegmentation}. Specifically, our approach achieved a 1.1\% improvement in mIoU on the DSD and a 1.3\% improvement on the Oralvision dataset, compared to the HQ-SAM model. These results highlight the superior segmentation quality offered by our approach over SAM-based approaches. Furthermore, when compared to other linear mechanism-based approaches, such as U-Mamba and SegMan, our algorithm not only outperformed these approaches in terms of both mIoU and mBIoU but did so with significant improvements, underscoring the effectiveness of our approach in providing high-quality segmentation results for dental images.

\begin{figure}[htbp]
    \centering
    \includegraphics[width=0.75 \linewidth]{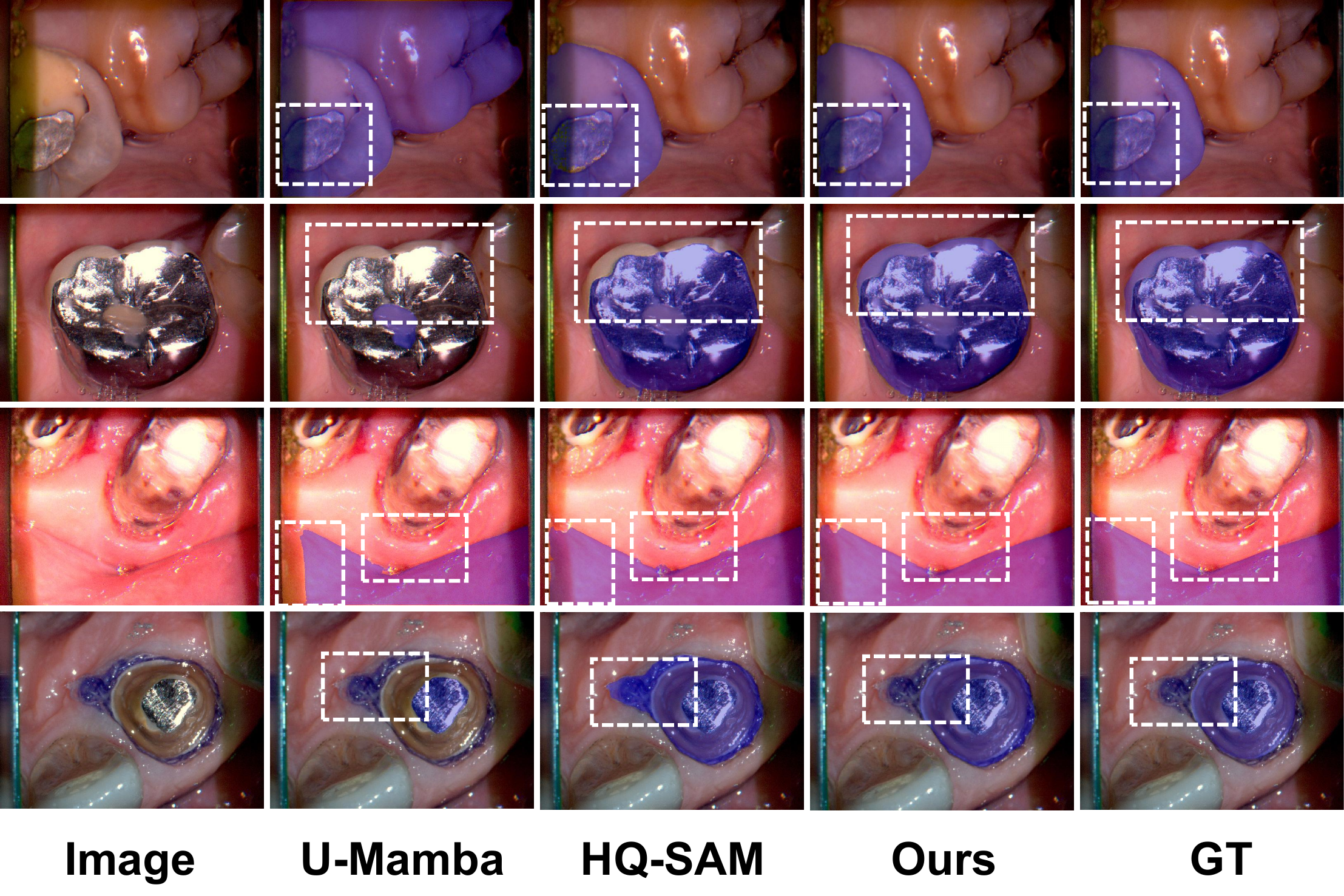}
    \caption{\textbf{Visualization results of U-Mamba, HQ-SAM, and our approach on DSD.} Our approach generates finer segmentation masks compared to others. The blue area shows the segmentation mask, and the white dashed box highlights differences between our approach and the others.}
    \label{figCompare}
    \vspace{-2mm}
\end{figure}

We conducted a qualitative comparison with U-Mamba and HQ-SAM, and visualized the segmentation results on the DSD in Fig.~\ref{figCompare}. As shown, our approach demonstrates a greater ability to adapt to complex oral environments, thereby producing more fine-grained segmentation results.

During the acquisition of oral images, shadow occlusion is often unavoidable, leading to uneven illumination and degraded segmentation performance. In the first example, the segmentation result is adversely affected by both lighting inconsistencies and the presence of metal inlays. Consequently, U-Mamba fails to completely segment the patient's molars, while HQ-SAM produces a result with numerous holes. In contrast, our approach not only successfully segments the entire molar region but also captures more fine details, enabling it to better resist such interference and generate masks with fewer holes. Moreover, due to the high visual similarity between the cheek and the gum, previous approaches often struggle to distinguish the boundary, resulting in incorrect segmentation. However, in the third example, our approach successfully differentiates between the gum and cheek regions, effectively suppressing noise from shadows and reflections, and thus achieves a more refined segmentation outcome.

\begin{table}[htbp]
\centering
\small
\caption{Complexity comparison of different segmentation methods.}
\label{complexity}
\begin{tabular}{lccc}
\toprule
Approach & Params (M) & FLOPs (G) & Latency (ms) \\
\midrule
Vim          & 26.1 & 41.3 & 20.7 \\
U-Mamba      & 32.8 & 48.6 & 24.9 \\
HQ-SAM       & 91.4 & 168.2 & 69.5 \\
FU-Mamba (Ours) & 36.2 
                  & 53.1 
                  & 28.2 \\
\bottomrule
\end{tabular}
\end{table}

Although FU-Mamba introduces additional modules such as the Offset Prediction Network and frequency-domain enhancement operations, the computational overhead remains within a reasonable range. As shown in Table \ref{complexity}, FU-Mamba achieves a favorable trade-off between segmentation accuracy and computational efficiency. Compared with U-Mamba, the proposed approach introduces only moderate increases in parameters, FLOPs, and inference latency, while still maintaining substantially lower complexity than foundation-based models such as HQ-SAM. Specifically, FU-Mamba achieves over 2\% improvement in mIoU on the DSD dataset with only 4.5G additional FLOPs and 3.3 ms extra inference latency.

\subsection{Discussion}
Dental image segmentation plays a vital role in the digital transformation of clinical oral diagnosis and treatment. The proposed FU-Mamba segmentation approach builds upon the UNet architecture by integrating the advantages of a dynamic scanning mechanism and a frequency-balancing strategy, thereby enabling the generation of high-quality segmentation results for dental images.

The experimental findings derived from both the dental segmentation dataset and the Oralvision dataset illustrate the efficacy of the proposed methodology. In particular, the dynamic scanning mechanism is shown to more effectively capture local information, which in turn improves the model's adaptability to diverse image content. Furthermore, by balancing high- and low-frequency features, our approach improves dental segmentation performance in challenging oral environments, such as those affected by high-frequency noise and poor lighting conditions.

The DMB in FU-Mamba demonstrates superior performance in dental image segmentation, primarily due to its effective modeling of the spatial continuity inherent in oral anatomical structures. By adaptively predicting sampling offsets, the model reorganizes image patches in a structure-aware manner that better aligns with the semantic layout of the image. This optimized sequence enhances the state space model's capacity to learn spatial dependencies among various regions, thereby improving its ability to delineate boundaries between teeth and adjacent tissues. As a result, this approach markedly advances the modeling of essential structures and contributes to the production of high-quality, detailed segmentation masks.

The frequency domain enhancement block demonstrates notable advantages under challenging conditions such as low illumination or imaging noise, making it a critical component in enhancing the segmentation performance of FU-Mamba. Specifically, FEB integrates wavelet decomposition with frequency-aware pooling to amplify mid-frequency features that carry structural semantics while simultaneously achieving a balanced representation between local details and global structures. Consequently, the module enables the model to retain essential shape information and boundary cues even when fine image details are degraded.

\begin{figure}[htbp]
    \centering
    \includegraphics[width=0.6 \linewidth]{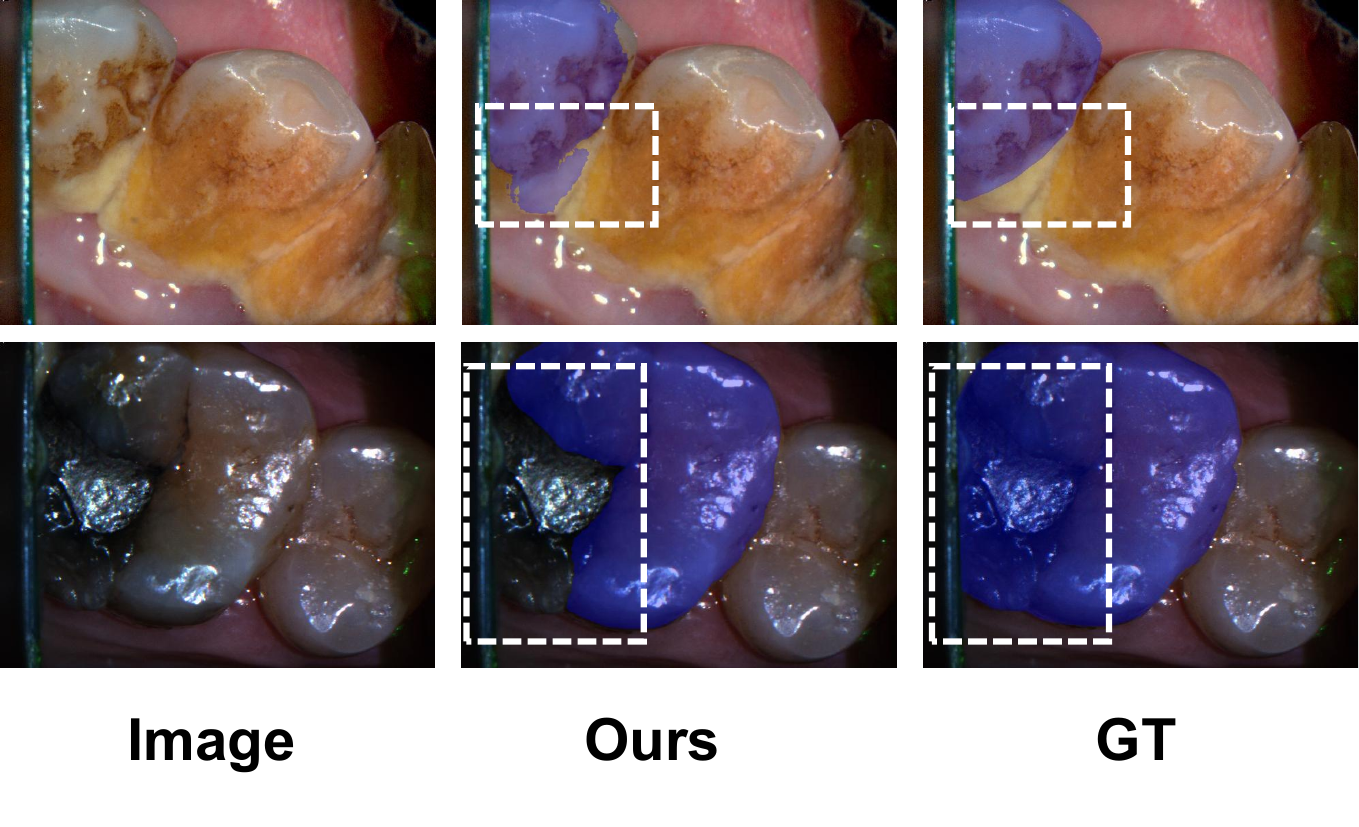}
    \caption{\textbf{Failure cases of our approach in DSD.} The blue regions represent the masks, and the white dashed boxes highlight segmentation differences.}
    \label{figDrawback}
\end{figure}
As illustrated in Fig.~\ref{figDrawback}, although the proposed method achieves robust performance on most dental images, failure cases still occur under extremely challenging conditions, such as severe noise, strong reflections, low contrast, and heavy dental calculus occlusion. To further evaluate robustness, we additionally analyzed the failure distribution on the DSD test set and observed that approximately 6.5\% of the samples showed noticeable segmentation degradation, mainly involving boundary leakage and local structure omission.
From a module-level perspective, these failures are primarily related to the limitations of dynamic offset estimation and frequency-domain enhancement under adverse imaging conditions. In the DMB module, unreliable local features may cause the Offset Prediction Network to generate unstable sampling offsets, leading to inaccurate structural aggregation during SSM modeling. Meanwhile, although the Frequency Enhancement Block enhances discriminative frequency components, its effectiveness may decrease when high-frequency noise and reflection artifacts dominate the spectrum, making boundary-related information difficult to distinguish from interference signals.
These observations suggest that improving the robustness of dynamic sampling and adaptive frequency modeling under extreme imaging conditions remains an important direction for future research.

Future work will focus on addressing the current limitations by incorporating multimodal data and enhancing frequency-aware processing. Specifically, introducing additional modalities such as depth or near-infrared images may provide complementary structural information, thereby improving robustness in occluded or low-contrast regions. Moreover, refining the frequency domain enhancement block to adaptively adjust the balance between high- and low-frequency components based on local context may further suppress noise and improve boundary precision.

\section{Conclusion}\label{conclusion}
This paper presents FU-Mamba, a dental image segmentation approach that integrates a dynamic scanning mechanism with a frequency domain enhancement block. The experimental findings indicate that the proposed methodology attains competitive segmentation performance across two dental datasets. Specifically, the DMB generates reordered sequences based on adaptively adjusted positions, thereby enabling more accurate capture of critical structures during segmentation. Concurrently, the FEB facilitates the equilibrium of spectral components, thereby augmenting the model's ability to accurately depict intricate frequency characteristics. However, the model still exhibits limitations in capturing fine-grained structures under noisy conditions, which may result in incomplete or inaccurate boundary delineation. Future work will focus on incorporating multi-modal data and improving frequency-aware modeling to further enhance segmentation robustness and accuracy.

\section*{Declaration of Competing Interest}
All authors declare that they have no conflicts of interest with respect to the publication of this article.

\section*{Availability of Data and Materials} 
Data supporting the findings of this study are available from the corresponding author on a reasonable request.

\section*{Acknowledgements}
This work was supported by the Zhejiang Province Natural Science Foundation (No. LZ24F020001), the Open Project Program of State Key Laboratory of Advanced Medical Materials and Devices (No. SQ2022SKL01089-2025-14), and the Open Project Program of State Key Laboratory of Virtual Reality Technology and Systems, Beihang University (No. VRLAB2026B13).

\bibliographystyle{elsarticle-num}
\bibliography{mybibfile2}

\end{document}